\pdfoutput=1

\documentclass[11pt]{article}

\usepackage[final]{acl}

\usepackage{times}
\usepackage{latexsym}

\usepackage[T1]{fontenc}

\usepackage[utf8]{inputenc}

\usepackage{microtype}

\usepackage{inconsolata}

\usepackage{graphicx}

\usepackage{tcolorbox}
\usepackage{enumitem}

\title{
Scaling Laws Are Unreliable for Downstream Tasks:
A Reality Check
}

\author{
  Nicholas Lourie$^{1}$\thanks{Equal contribution. Code: \url{https://github.com/nicholaslourie/scale-fails}} \quad
  Michael Y. Hu$^{1}$\footnotemark[1] \quad
  Kyunghyun Cho$^{1,2,3}$ \\
  $^{1}$New York University \quad $^{2}$Prescient Design \quad $^{3}$ CIFAR LMB \\
  \texttt{\{nick.lourie, michael.hu, kyunghyun.cho\}@nyu.edu}
}

\begin{document}

\maketitle

\begin{abstract}
Downstream scaling laws aim to predict task performance at larger scales from the model's performance at smaller scales. Whether such prediction should be possible is unclear: some works discover clear linear scaling trends after simple transformations of the performance metric, whereas others point out fundamental challenges to downstream scaling laws, such as emergence and inverse scaling. In this work, we conduct a meta-analysis of existing data on downstream scaling laws, and we find that predictable scaling only occurs in a minority of cases: 39\% of the time. Moreover, seemingly benign changes to the experimental setting can completely change the scaling behavior. Our analysis underscores the need to understand the conditions under which scaling laws succeed. To accurately model the relationship between pretraining loss and task performance, we must embrace the cases in which scaling behavior deviates from linear trends.
\end{abstract}

\section{Introduction}
\label{sec:introduction}

Scaling laws for pretraining establish that the loss improves reliably when increasing the size of the model, training data, or compute \citep{kaplan2020,hoffmann2022an,pearce2024reconciling}. However, better pretraining loss does not always translate to better downstream performance \citep{magnusson2024paloma}. This gap can be caused by a variety of issues; among the best known is \textit{emergence}, or the fact that on some tasks models below a certain scale show no trend or near-chance performance \citep{wei2022emergent}. In addition, model performance can increase then decrease, in what is called \textit{inverse scaling} \citep{mckenzie2023inverse,wilcox2024bigger}. Despite these challenges, several works also find downstream performance is roughly linear in the pretraining loss, possibly after a transformation \citep{huang2024compression, gadre2025language, chen2025scaling}.

\begin{figure}[t]
    \centering
    \includegraphics[width=\linewidth]{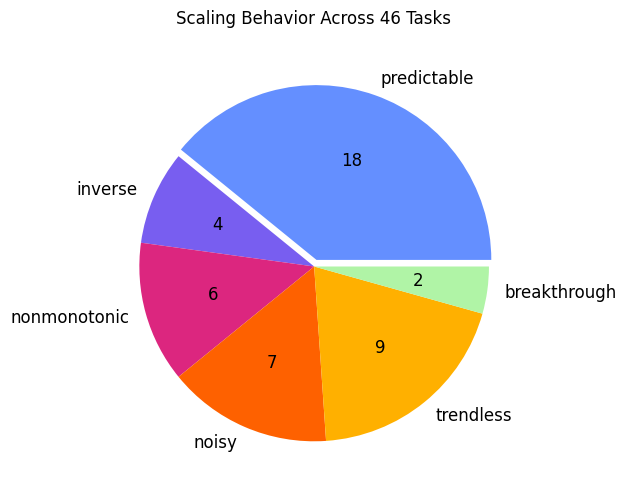}
    \caption{
        Revisiting the 46 tasks studied in \citet{gadre2025language}, we find that only 18 tasks---or 39\%---demonstrate smooth, predictable improvement (Figure \ref{fig:scaling-behaviors_predictable}). The other 28 tasks are shown in Figures~\ref{fig:scaling-behaviors_inverse} through \ref{fig:scaling-behaviors_breakthrough}, where we group them into different degenerate scaling behaviors: inverse, nonmonotonic, noisy, trendless, and breakthrough scaling. See Figure~\ref{fig:scaling-behaviors} for examples.
    }
    \label{fig:pie-chart}
\end{figure}

\begin{figure*}[t!]
    \centering
    \includegraphics[width=\linewidth]{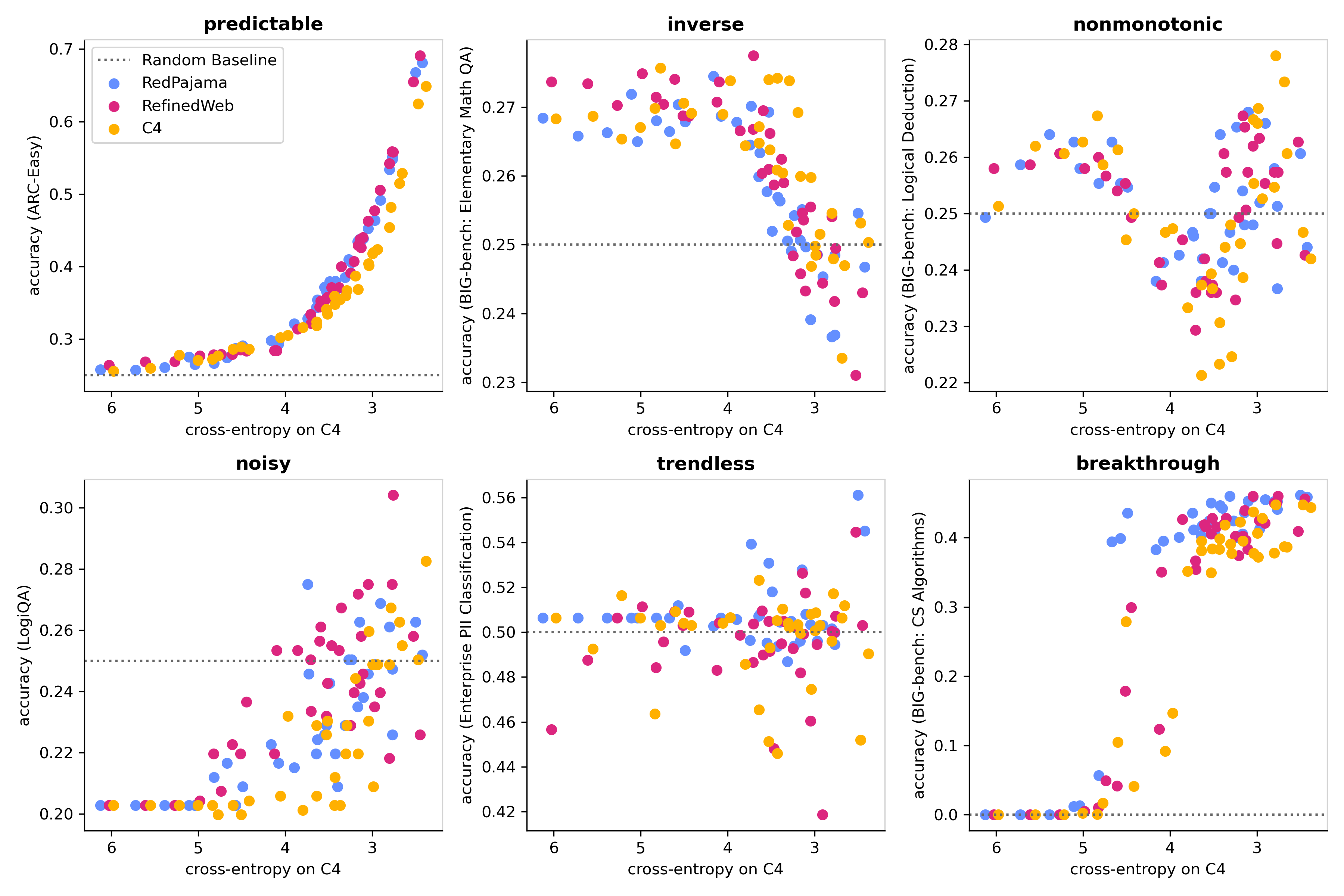}
    \caption{
        A taxonomy of different scaling behaviors. \textbf{Predictable} scaling fits closely to a linear functional form after, for example, exponentiating the cross-entropy loss. However, depending on the downstream task, models do not always improve with scale (\textbf{inverse}, \textbf{nonmonotonic}, and \textbf{trendless}), or the improvement might be highly \textbf{noisy}. The improvement might also follow a functional form that is difficult to extrapolate like a sigmoid (\textbf{breakthrough}).
    }
    \label{fig:scaling-behaviors}
\end{figure*}

Thus, the extent to which downstream scaling laws work is unclear. How can they follow linear forms when we know of many tasks that exhibit emergence or inverse scaling? Are these just edge cases, or are they common, with no established explanation? Here, we aim to clarify this confusion. We explore and consider three core factors affecting downstream scaling laws: 1) the data used for pretraining and validation, 2) the downstream task, and 3) the experimental setup. Realistic changes to each of these factors can change the relationship to downstream performance, to the point that the scaling law's functional form might no longer hold. The behavior can even change qualitatively---with a decrease in trend under one set of conditions flipping to an increase under another. Such qualitative changes can not be removed by a new functional form. They pose fundamental obstacles to studying large scale models with small scale proxies.

In particular, we find that:
\begin{enumerate}
    \item Choosing a different dataset for the validation perplexity can flip scaling trends. For instance, models pretrained on one corpus might appear to improve faster than another, but this trend can reverse with different validation data (\S \ref{sec:payn}).
    \item Revisiting a prior study \citep{gadre2025language}, we only find predictable scaling in a minority of cases: 39\%. Phenomena like emergence or inverse scaling can actually be quite common, occurring for many downstream tasks (\S \ref{sec:base-rates}).
    \item Scaling behavior from one experimental setup can \textit{qualitatively} change under another; a task with predictable scaling could become non-monotonic or even show no trend at all (\S \ref{sec:experimental-setup}).
\end{enumerate}

Our analysis suggests that scaling laws are \textit{context-specific}. As such, we cannot assume downstream scaling will always be strictly linear. Rather, we need to better understand the failure modes of existing scaling laws and develop a holistic (and perhaps more complex) model of how foundation models improve on downstream tasks.

\section{Background}
\label{sec:background}

Downstream scaling laws try to extrapolate the performance of large-scale models from small-scale proxies. When successful, these scaling laws enable cost-effective experiments at the small scale that transfer to the large. Scaling laws for pretraining are well established \citep{rosenfeld2020a, kaplan2020, hoffmann2022an}; however, the ultimate goal of language models is to perform well on downstream tasks. As such, downstream scaling laws are also of great interest.

Unlike pretraining, there is little consensus on how to approach downstream scaling laws. Early efforts tried to predict downstream performance directly from parameters, data, or compute \citep{ivgi-etal-2022-scaling, Mahmood_2022_CVPR, openai2023gpt4}, but downstream performance often showed a noisy relationship to these quantities \citep{tay2022scale}. Other efforts sought surrogates for scale, like latent capabilities \citep{ruan2024observational} or task-specific losses \citep{grattafiori2024llama3herdmodels, bhagia2024establishingtaskscalinglaws}, in the hope of mapping from compute to surrogate, and from surrogate to downstream performance.

Many works have found downstream scaling laws are more stable when stated in terms of pretraining loss, the most widely used surrogate for scale \citep{xia-etal-2023-training, huang2024compression, du-etal-2024-understanding, gadre2025language, chen2025scaling}. Specifically, if two models differ in their number of parameters or pretraining tokens but still attain the same pretraining loss, then they tend to achieve the same downstream performance \citep{xia-etal-2023-training, du-etal-2024-understanding, gadre2025language}. Some authors generalize this principle, viewing pretraining loss as a general way to compare models---not just with different scales but different training recipes \citep{huang2024compression, chen2025scaling}. However, while pretraining loss correlates with downstream performance to a surprising degree, this relationship is far from absolute \citep{tay2022scale}.

In the best case, downstream task performance is roughly linear in some monotonic transformation of validation cross-entropy loss ($x$) and downstream metric ($y$):
$$ f(y) = a\, g(x) + b $$
For instance, \citet{gadre2025language} relate error rate  $y$ to the validation loss $x$ via: $y = \epsilon - k\, \exp\{-\gamma\,x\}$. The functions, $f$ and $g$, must depend on the metric; in some cases, a linear relationship alone could be enough \citep{huang2024compression, chen2025scaling}.

Such global structure enables extrapolation---the ultimate goal of downstream scaling laws; however, predictable structure might not exist. Phenomena such as emergence, inverse, and U-shaped scaling preclude this kind of global structure \citep{wei2022emergent, mckenzie2023inverse, wei-etal-2023-inverse}. They destroy it by creating \textit{structural breaks}, or points where the function describing one part of the scaling curve does not describe another \citep{andrews1993tests}. Without global structure, it is impossible to extrapolate from small to large. Certain tasks, like multiple choice questions, are more susceptible to emergence \citep{schaeffer2025why}. Sometimes emergence can be mitigated by choosing a different downstream metric \citep{schaeffer2023mirage}, but in other cases breakthrough improvements remain ``stubbornly emergent'' \citep{zhao2025distributionalscalinglawsemergent, du-etal-2024-understanding}. \citet{li2025misfitting} found scaling laws generally difficult to reproduce and sensitive to the functional form, training setup, data collection, and fitting algorithm, while \citet{magnusson2025datadecidepredictbestpretraining} found that comparing models at a smaller scale---without extrapolation---performed as well or better than scaling laws in determining the best data mix.

\begin{figure}[ht]
    \centering
    \includegraphics[width=\linewidth]{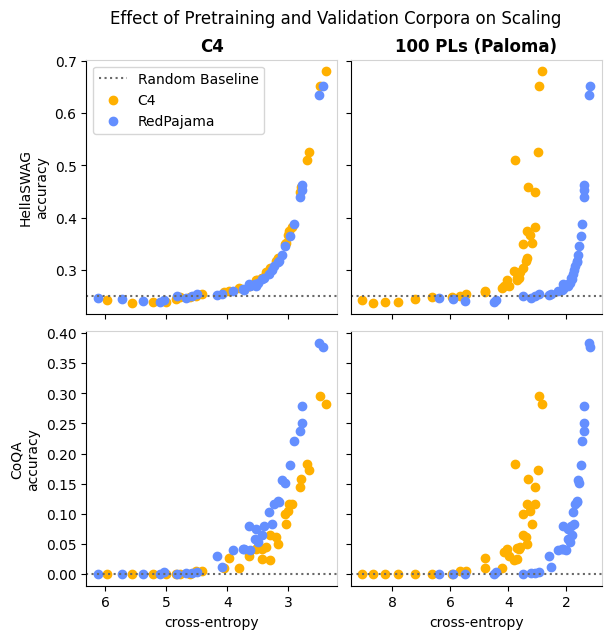}
    \caption{Choosing a different validation corpus can exaggerate or even reverse which pretraining corpus appears superior. On HellaSwag, the C4 corpus seems better than RedPajama when using 100 PLs as the validation set. Conversely, the scaling trends on CoQA for C4 and RedPajama flip when computing validation perplexity on C4 versus 100 PLs.}
    \label{fig:effect-of-validation-corpus}
\end{figure}

\section{Scaling Laws Are Specific to the Data}
\label{sec:payn}

Downstream scaling laws depend on several factors, including the pretraining data, validation data, and downstream task. If you vary the pretraining data, then you must fix a validation corpus to compare the loss across models. Once it is fixed, you might hope to focus on the validation loss alone and simplify your research. It is unclear how to choose the correct validation data but, what is more, its choice can entirely reverse which pretraining setup appears superior for a downstream task if we do not consider the full context. The pretraining data, validation data, and downstream task all interact in forming the scaling law---you can not consider one without the other two. In particular, you can not compare data mixes by their validation losses or downstream scaling laws based upon them.

To evince these claims, we reexamine the results from \citet{gadre2025language}, who pretrained models over different corpora. We observe how pretraining data, validation data, and task all interact in forming the scaling law (Figure~\ref{fig:effect-of-validation-corpus}). The colors correspond to pretraining corpora: C4 \citep{2020t5} and RedPajama \citep{weber2024redpajama}; the columns correspond to validation data: C4 \citep{2020t5} and Paloma's 100 Programming Languages \citep{magnusson2024paloma}; and the rows correspond to tasks: CoQA \citep{reddy-etal-2019-coqa} and HellaSwag \citep{zellers-etal-2019-HellaSwag}. If you ignore the other factors, then choice of validation data can completely change the scaling trend:

\paragraph{Exaggerating differences.} For HellaSwag with C4 as the validation corpus (top left), pretraining on either C4 or RedPajama produces the same scaling law. However, when using 100 Programming Languages (100 PLs, top right), the scaling laws for C4 and RedPajama no longer superimpose---pretraining on C4 appears to achieve much better performance even for a worse validation loss.

\paragraph{Flipping scaling trends.} Changing the task to CoQA (bottom left) also changes the scaling laws, with RedPajama now achieving better performance sooner. Even worse, changing the validation corpus from C4 to 100 PLs reverses this relationship \textit{again} (bottom right).

\paragraph{} Thus, whether or not better perplexity translates to better downstream performance depends on the task, the pretraining corpus, and the validation loss. Changes to any one of these three factors can reverse which pretraining setup appears superior.

\begin{figure*}[ht]
    \centering
    \includegraphics[width=\linewidth]{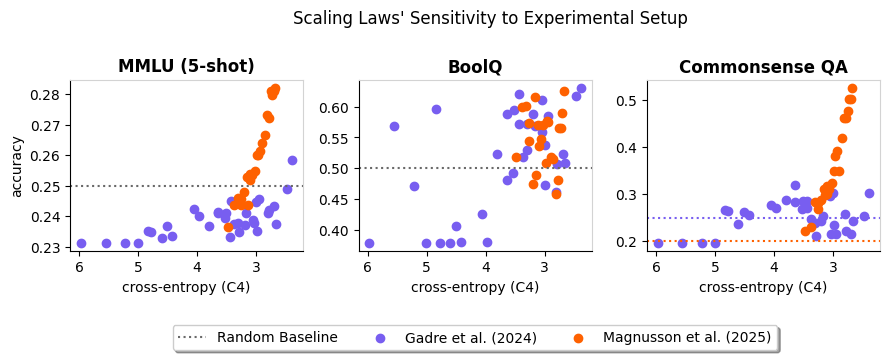}
    \caption{
        Scaling behavior changes depending on the experimental setting. \citet{gadre2025language} and \citet{magnusson2025datadecidepredictbestpretraining} both train language models on C4 and evaluate on MMLU, BoolQ, and Commonsense QA. Still, they differ in their details, such as model architecture, task formatting, or the number of answer choices (in the case of Commonsense QA). \textit{Even with the same corpora and downstream task, scaling trends can be dramatically different.}
    }
    \label{fig:scaling-changes}
\end{figure*}

\section{Irregular Scaling Is Common}
\label{sec:base-rates}

Phenomena like emergence and inverse scaling suggest that linear scaling laws do not capture all task scaling behavior. However, it is also unclear how prevalent these phenomena are in practice. Thus, we re-examine scaling behavior on the 46 tasks tested by \citet{gadre2025language}, classifying them into six categories visualized in Figure~\ref{fig:scaling-behaviors}. We find that linear scaling actually occurs in a \textit{minority} of cases in their setting: 39\% of the time (Figure~\ref{fig:pie-chart}). For some experimental setups, non-linear scaling is actually the norm. Are the experimental choices of \citet{gadre2025language} abnormal? On the contrary, all three of pretraining corpora, validation datasets, and downstream tasks are from well-known sources in the literature. Downstream tasks are comprised of established evaluations like BoolQ, HellaSwag, and BIG-Bench \citep{clark-etal-2019-boolq,zellers-etal-2019-HellaSwag,srivastava2023beyond}. Irregular scaling occurs within popular tasks and is easy to find.

\section{Scaling Behavior Is Not Always Robust}
\label{sec:experimental-setup}

Finally, we show that conclusions about scaling laws may not generalize across settings: setups with the same validation data and downstream tasks may observe entirely different scaling trends. To show this, we take the 10 overlapping tasks between \citet{gadre2025language} and \citet{magnusson2025datadecidepredictbestpretraining}. These authors consider several of the same pretraining corpora and downstream tasks; however, their implementation details differ (see Appendix~\ref{app:data-sources}). For example, \citet{gadre2025language} use fewer answer choices for Commonsense QA. Thus, we might expect to see quantitative differences in some of their scaling laws, but what is more surprising is that we also see \textit{qualitative} changes in their scaling behavior.

To enable comparison, we evaluate all models from \citet{gadre2025language} and \citet{magnusson2025datadecidepredictbestpretraining} on the same validation corpus (C4). For \citet{magnusson2025datadecidepredictbestpretraining}, which does not use C4, we evaluate 200 released checkpoints, which vary in their pretraining setup:\footnote{The 20M models are the smallest models in \citet{magnusson2025datadecidepredictbestpretraining} and have the highest cross-entropy loss.}
\begin{enumerate}[itemsep=0pt, parsep=0pt, topsep=3pt]
    \item Parameters: \{20M, 60M, 150M, 300M, 1B\}.
    \item Pretraining corpus: \{Dolma, C4, DCLM-Baseline, RefinedWeb (Falcon), FineWeb-Pro\} \citep{soldaini-etal-2024-dolma,2020t5,li2024datacomplm,refinedweb,zhou2025programming}.
    \item Training steps: from 5,000 to 40,000, in intervals of 5,000.
\end{enumerate}

Out of the 10 overlapping downstream tasks, Figure~\ref{fig:scaling-changes} shows three that produce different scaling trends between setups. In both setups, MMLU shows a positive trend; however, the noisiness and shape of that trend differs greatly. At an even greater extreme, CommonsenseQA's scaling behavior qualitatively changes: while CommonsenseQA shows nonmonotonic scaling in \citeposs{gadre2025language} results, it exhibits a clean scaling law with \citeposs{magnusson2025datadecidepredictbestpretraining}. On the other hand, in \citeposs{magnusson2025datadecidepredictbestpretraining} results, BoolQ's scaling law appears trendless; however, this lack of trend comes from the models spanning too small a validation loss range. Since \citeposs{gadre2025language} models cover a wider range of validation losses, the trend more clearly emerges. Thus, scaling behavior can fluctuate, even between controlled studies for the same task with the same pretraining corpora.

\section{Discussion}
\label{sec:discussion}

Given the cost of training modern foundation models, scaling laws have become an invaluable tool for making informed modeling decisions. Scaling laws enable us to extrapolate results where compute costs would otherwise make extensive experimentation infeasible. However, extrapolations are only worthwhile when their assumptions are faithful to the data.

As practitioners of scaling laws, we must realize that predictable scaling laws often exist, but one cannot assume that they hold for all contexts. Even if scaling is stable on the same task for the same validation data, other aspects of the experimental setup might change the scaling behavior (\S\ref{sec:experimental-setup}). To some extent, scaling laws are \textit{investigator-specific} \citep{li2025misfitting}, and so each investigator must verify the scaling law's presence with visualizations and regression diagnostics \citep{shalizi2015truth}.

For researchers, downstream scaling laws offer many fascinating new directions. Empirically, we need better ways to stabilize scaling laws and detect when irregular scaling might occur. We must understand the factors affecting scaling laws, and what parts of the experimental setup must remain constant for linear scaling laws to hold. Theoretically, we need a model for why predictable scaling occurs \citep{hutter2021learning}, and a core goal of such a theory should be explaining exactly the cases in which it does not.

\section{Conclusion}
\label{sec:conclusion}

In this work, we surveyed where downstream scaling laws break down. Depending on the pretraining corpus, validation corpus, or downstream task, scaling laws can change. Better perplexity does not always translate to better downstream performance (\S\ref{sec:payn}); \textit{perplexity is not all you need}.
Even when holding pretraining and validation data the same, more often than not a predictable scaling law does not exist at all (\S\ref{sec:base-rates}). Irregular behaviors like nonmonotonic, trendless, or breakthrough scaling are all common; one must establish predictable scaling for the given task before relying on it. Finally, seeing predictable scaling in one experimental setup does not guarantee it for another (\S\ref{sec:experimental-setup}). Until we better understand why predictable scaling arises and its sufficient conditions, investigators must verify scaling laws in their own settings.

\section*{Limitations}
\label{sec:limitations}

Our work uses data and model checkpoints from existing studies \citep{gadre2025language,magnusson2025datadecidepredictbestpretraining}. While this is sufficient for the counterexamples featured in this work, there may be unknown biases shared between these two projects that we have missed.

In addition, our study establishes that downstream scaling laws are unreliable \textit{under current practice}. Neural networks used to be notoriously difficult to train; however, as understanding developed around how to architect, initialize, and optimize them, training neural networks became routine and reliable. In a similar way, it is possible that the difficulties discussed here may be overcome by better techniques for measuring and estimating scaling laws. We hope the challenges we identify here inspire researchers to search for them.

\section*{Acknowledgments}
\label{sec:acknowledgments}

MYH is supported by the NSF Graduate Research Fellowship. This work was supported by the Institute of Information and Communications Technology Planning and Evaluation (IITP) with a grant funded by the Ministry of Science and ICT (MSIT) of the Republic of Korea in connection with the Global AI Frontier Lab International Collaborative Research. This work was also supported by the Samsung Advanced Institute of Technology (under the project Next Generation Deep Learning: From Pattern Recognition to AI) and the National Science Foundation (under NSF Award 1922658). This work was supported in part through the NYU IT High Performance Computing resources, services, and staff expertise.

\bibliography{anthology,custom}

\clearpage
\appendix

\section{Data Sources}
\label{app:data-sources}

The first set of results comes from \citet{gadre2025language} and is available\footnote{\url{https://github.com/mlfoundations/scaling}} under the MIT License. \citet{gadre2025language} pretrain transformer language models across different scales on several different corpora (e.g., C4 \citep{2020t5}, RedPajama \citep{weber2024redpajama}, and RefinedWeb \citep{penedo-etal-2023-refinedweb}). For each model, they compute its validation loss on these and other corpora, and evaluate the model with few-shot prompting across 46 tasks using LLM Foundry.\footnote{\url{https://github.com/mosaicml/llm-foundry}}

The second set of results comes from \citet{magnusson2025datadecidepredictbestpretraining} and is available\footnote{\url{https://huggingface.co/datasets/allenai/DataDecide-eval-results}} under the ODC-By License. \citet{magnusson2025datadecidepredictbestpretraining} also pretrain transformer language models but they choose different architectural details than \citet{gadre2025language} and use the C4 and RefinedWeb versions from Dolma 1.7 \citep{soldaini-etal-2024-dolma}. They pretrain models across these and other corpora and evaluate them with few-shot prompting on 10 different tasks via OLMES \citep{gu-etal-2025-olmes}.\footnote{\url{https://github.com/allenai/olmes}}

The evaluation harnesses, LLM Foundry and OLMES, have some important differences. LLM Foundry ships with its own versions of tasks' datasets. For some tasks (e.g., Commonsense QA and SIQA), it changes the number of answer choices, thus the random baseline can change between the different versions and we have indicated this in the appropriate figures. LLM Foundry also varies the number of shots depending on the task, whereas OLMES uses 5 curated examples for each one. Other differences include the task formulation (whether to use multiple-choice or cloze format) and implementation details such as the prompts.

Finally, LLM Foundry's version of SIQA had an error where the gold labels were incorrect.\footnote{\url{https://github.com/mosaicml/llm-foundry/pull/774}.} This issue was fixed in LLM Foundry v0.5.0;\footnote{\url{https://github.com/mosaicml/llm-foundry/releases/tag/v0.5.0}} however, \citet{gadre2025language} used v0.4.0 for at least some of their experiments.\footnote{\url{https://wandb.ai/samir/dcnlp/runs/rezso5ec/files/requirements.txt}} As a result, we do not examine SIQA in our analyses, although Figure~\ref{fig:sensitivity-to-experimental-setups_full} in Appendix~\ref{app:sensitivity-to-experimental-setups} includes SIQA for completeness.

\section{Reproducibility}
\label{app:reproducibility}

We augment the results from \citet{magnusson2025datadecidepredictbestpretraining} by evaluating their models on C4's validation data. This additional information enables us to compare scaling laws from \citet{magnusson2025datadecidepredictbestpretraining} and \citet{gadre2025language}.

To compute the validation loss, we ran the DataDecide \citep{magnusson2025datadecidepredictbestpretraining} models from HuggingFace using \texttt{ai2-olmo} (\url{https://github.com/allenai/OLMo}), computing the perplexity on C4's validation split using a batch size of 64. For inference, we used a combination of A100 and H100 GPUs with 32 GB of CPU RAM. Running inferences over the C4 validation set took approximately 10 minutes.

\section{Scaling Behaviors}
\label{app:scaling-behaviors}

Our analysis centers on \textit{qualitative} scaling behavior in order to avoid having its conclusions depend on a particular form of the scaling law. Without formal criteria, researchers might disagree on particular examples, but then researchers might also disagree on formal criteria. Thus we define the scaling behaviors informally, as follows:

\begin{description}[itemsep=0pt, parsep=0pt, topsep=3pt]
    \item[predictable] Performance increases, without too much variation around the trend.
    \item[inverse] Performance decreases as loss improves.
    \item[nonmonotonic] Performance switches between increasing and decreasing.
    \item[noisy] Performance increases but varies greatly around the trend.
    \item[trendless] Performance is flat or there is too much noise to discern a trend.
    \item[breakthrough] Performance starts flat, abruptly increases, and then plateaus.
\end{description}

\begin{figure*}
    \centering
    \includegraphics[width=\linewidth]{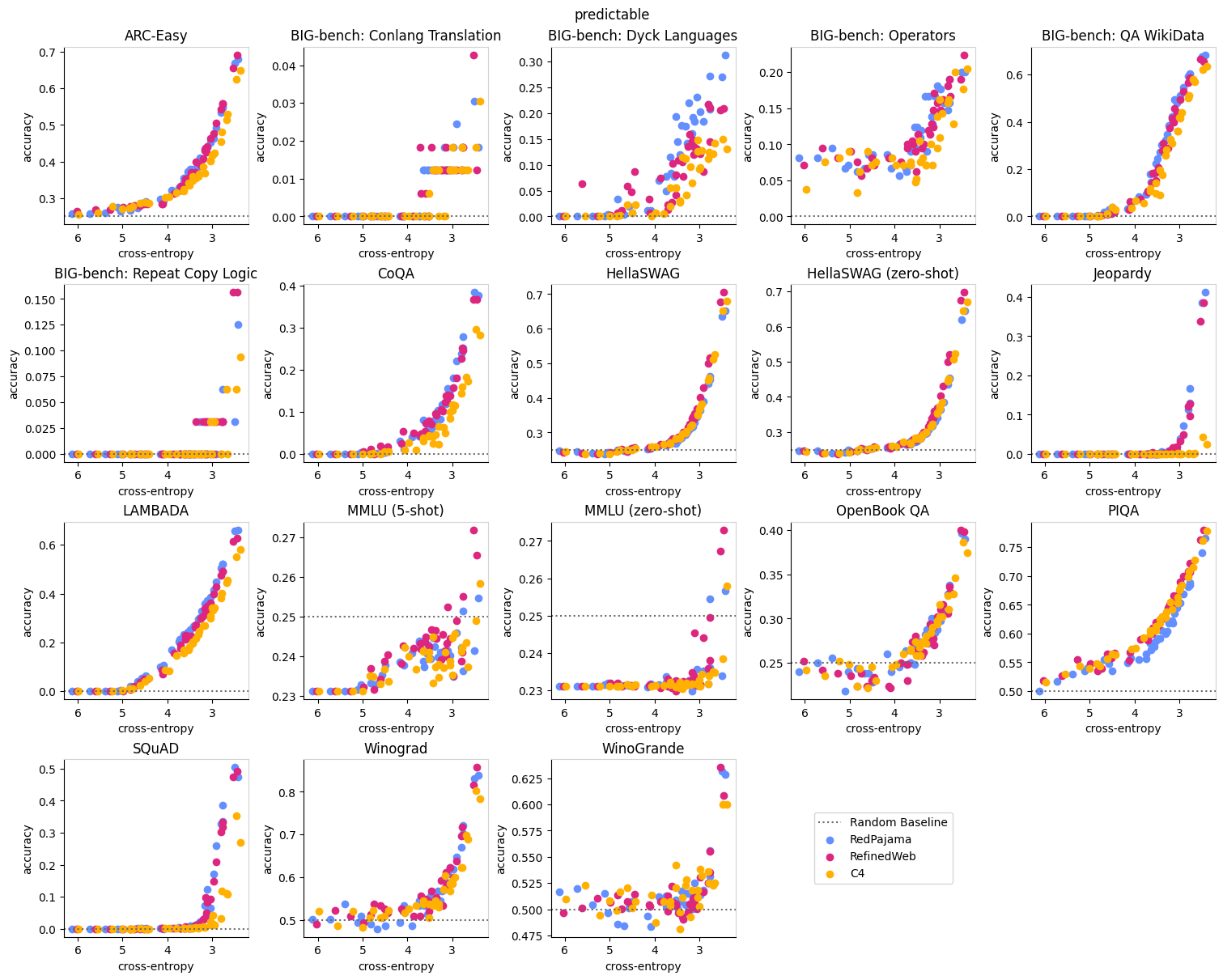}
    \caption{The 18 tasks in \citet{gadre2025language} with scaling behavior well-described by a linear scaling law after transforming the cross-entropy loss.}
    \label{fig:scaling-behaviors_predictable}
\end{figure*}

\begin{figure*}
    \centering
    \includegraphics[width=\linewidth]{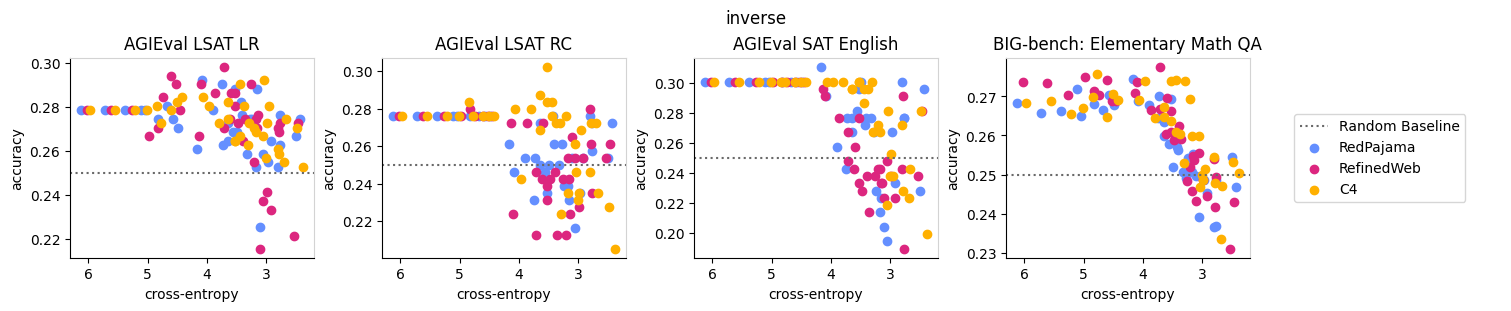}
    \caption{Tasks in \citet{gadre2025language} with inverse scaling.}
    \label{fig:scaling-behaviors_inverse}
\end{figure*}

\begin{figure*}
    \centering
    \includegraphics[width=\linewidth]{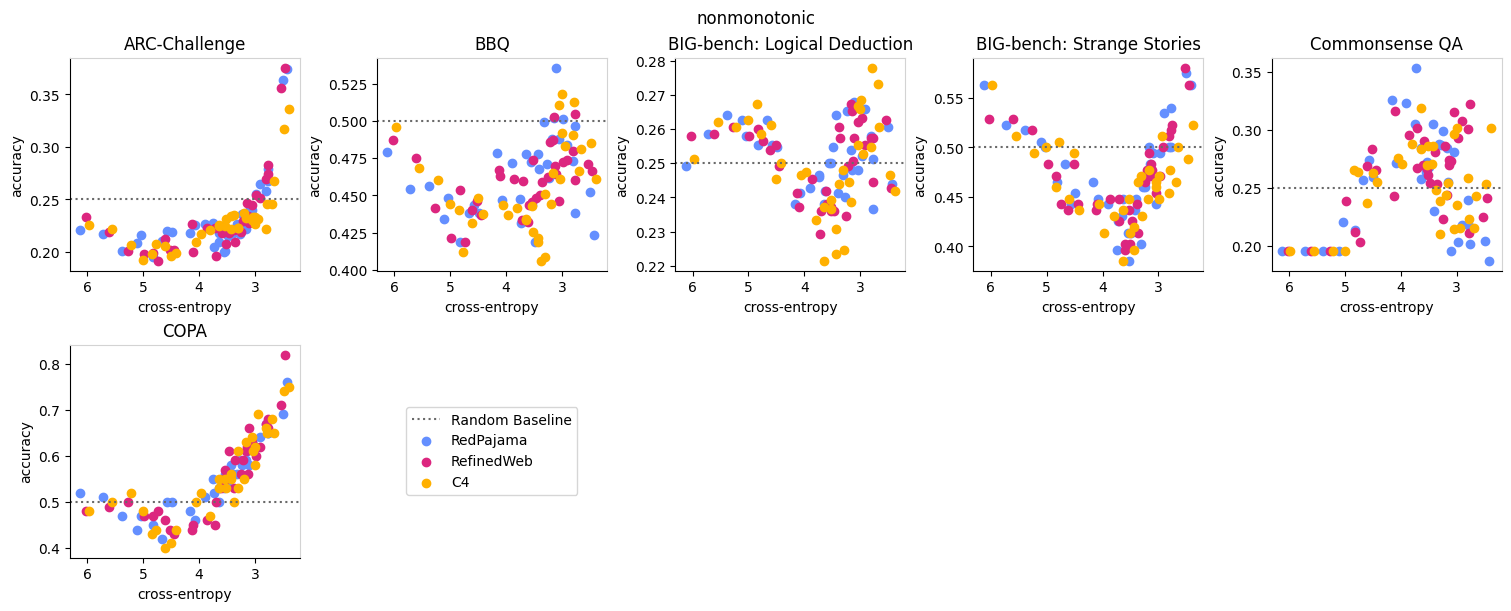}
    \caption{Tasks in \citet{gadre2025language} with nonmonotonic scaling.}
    \label{fig:scaling-behaviors_nonmonotonic}
\end{figure*}

\begin{figure*}
    \centering
    \includegraphics[width=\linewidth]{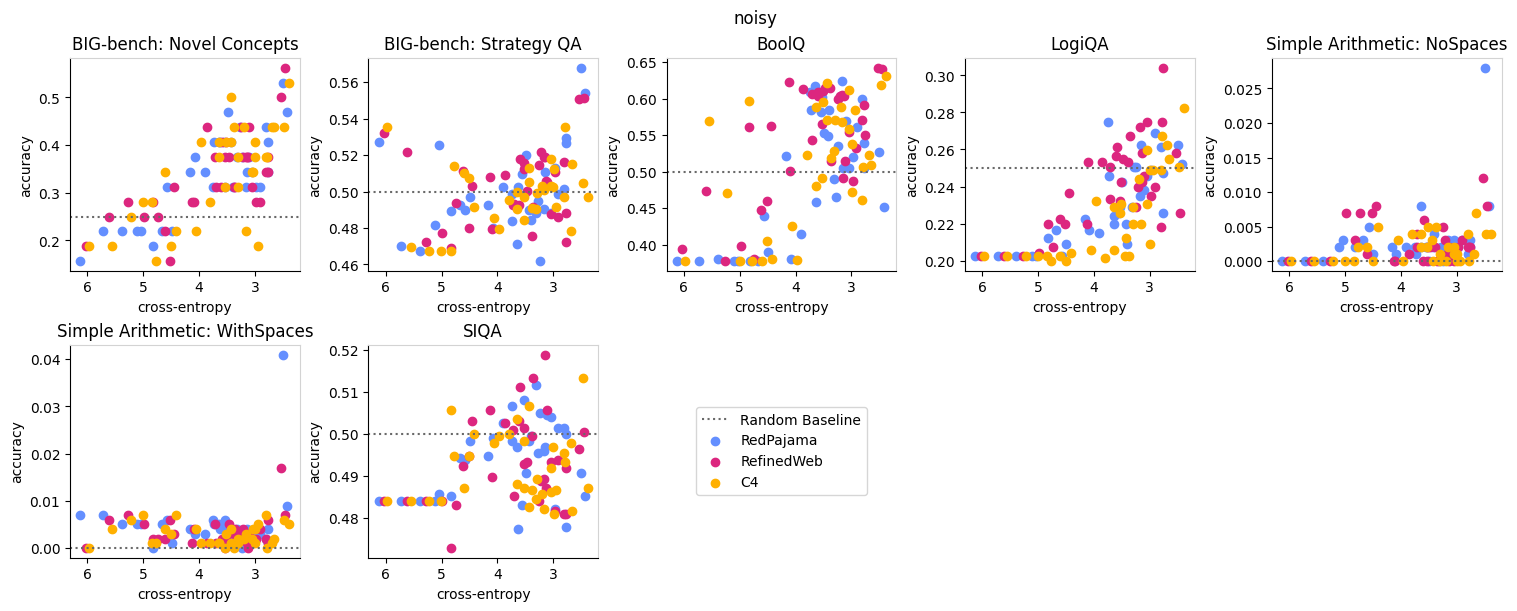}
    \caption{Tasks in \citet{gadre2025language} with noisy scaling.}
    \label{fig:scaling-behaviors_noisy}
\end{figure*}

\begin{figure*}
    \centering
    \includegraphics[width=\linewidth]{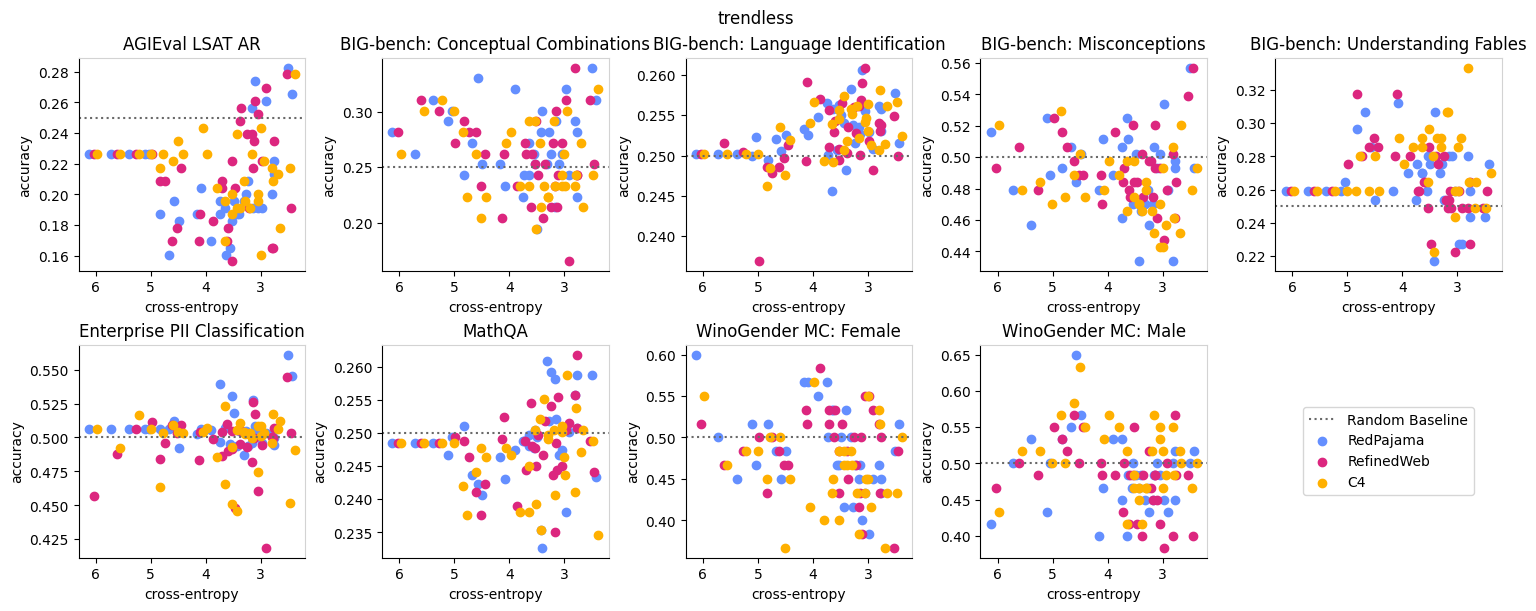}
    \caption{Tasks in \citet{gadre2025language} with no clear scaling trend.}
    \label{fig:scaling-behaviors_trendless}
\end{figure*}

\begin{figure*}
    \centering
    \includegraphics[width=\linewidth]{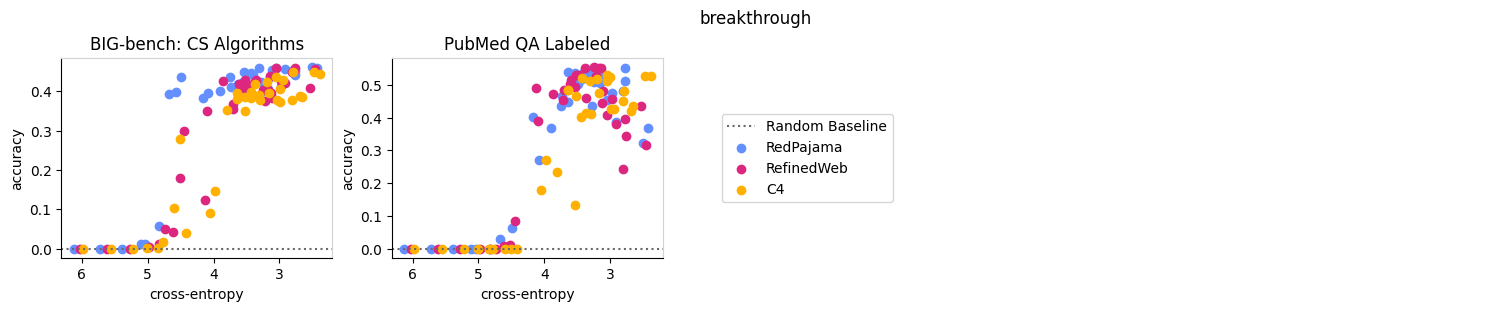}
    \caption{Tasks in \citet{gadre2025language} demonstrating breakthrough scaling.}
    \label{fig:scaling-behaviors_breakthrough}
\end{figure*}

Figures~\ref{fig:scaling-behaviors_predictable} through \ref{fig:scaling-behaviors_breakthrough} present the tasks studied in \citet{gadre2025language} sorted by scaling behavior. The figures include scaling laws fitted using the functional form from \citet{gadre2025language} that relates the error rate to the validation loss: $y = \epsilon - k\, \exp\{-\gamma\,x\}$. We apply this functional form to the accuracy using the fact that accuracy is one minus the error rate. The scaling laws were fitted by minimizing mean squared error with \texttt{differential\_evoluation} from SciPy \citep{scipy2020}. For the optimization, we used bounds of 0 to 1 for $\epsilon$, 0 to 20 for $\ln(k)$ (we searched $k$ on a log scale to widen the range), and 0 to 20 for $\gamma$. In addition, we used 60 for the \texttt{popsize} multiplier. Figures~\ref{fig:scaling-behaviors_predictable} through \ref{fig:scaling-behaviors_breakthrough} also include the $R^2$ of each scaling law's fit as a measure of how closely the scaling behavior adheres to the functional form.

\section{Sensitivity to Experimental Setups}
\label{app:sensitivity-to-experimental-setups}

\begin{figure*}
    \centering
    \includegraphics[width=\linewidth]{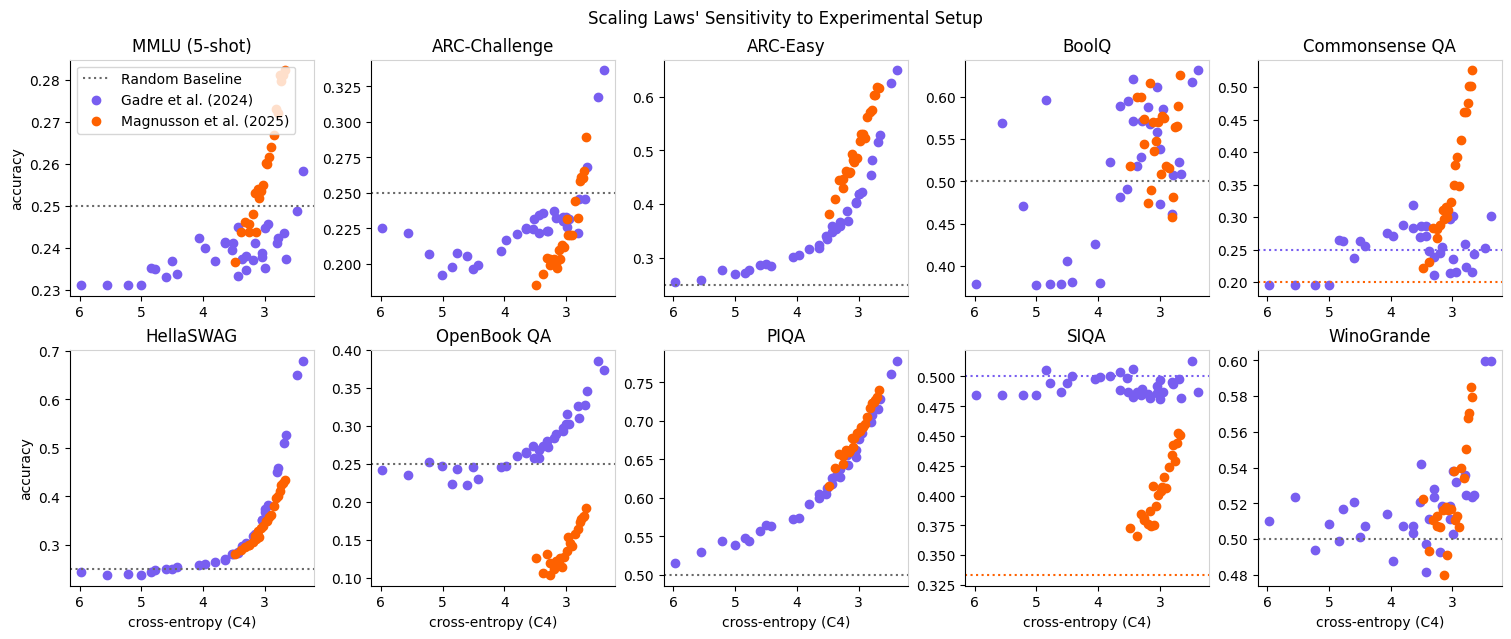}
    \caption{A comparison of scaling in the experimental setups of \citet{gadre2025language} and \citet{magnusson2025datadecidepredictbestpretraining}. Both trained language models on C4 and evaluated via few-shot prompting on the tasks above; however, their experimental setups differ: architectural details, prompts, number of shots, task format, and in some cases the number of answer choices (Commonsense QA and SIQA). Such experimental details totally change scaling behavior.}
    \label{fig:sensitivity-to-experimental-setups_full}
\end{figure*}

Figure~\ref{fig:sensitivity-to-experimental-setups_full} shows how the scaling laws change between the experimental setups of \citet{gadre2025language} and \citet{magnusson2025datadecidepredictbestpretraining} across 10 tasks.

\section{The Effect of the Pretraining and Validation Corpora}
\label{app:the-effect-of-the-pretraining-and-validation-corpora}

\begin{figure*}
    \centering
    \includegraphics[width=\linewidth]{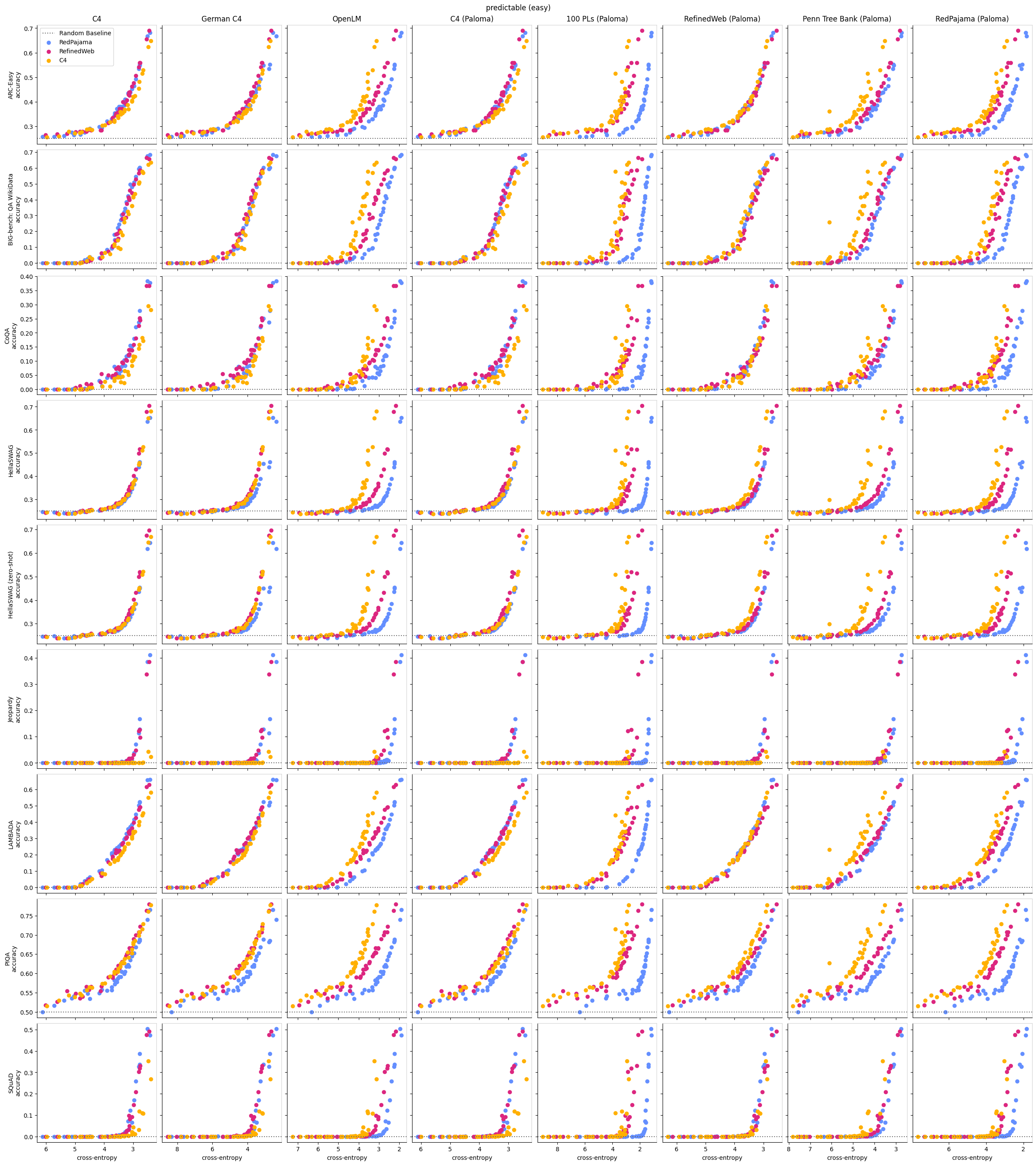}
    \caption{Different pretraining corpora will appear to be the best for downstream tasks, depending on choice of validation dataset.}
    \label{fig:effect-of-validation-corpus_predictable-easy}
\end{figure*}

Figure~\ref{fig:effect-of-validation-corpus_predictable-easy} compares how the task, pretraining, and validation corpora affect the scaling curve. For ease of visualization, we only show the effect on tasks with the cleanest, most predictable scaling laws.

\end{document}